\documentclass{article}

\usepackage[final,nonatbib]{neurips_2020_ml4ps}

\usepackage[utf8]{inputenc} 
\usepackage[T1]{fontenc}    
\usepackage{hyperref}       
\usepackage{url}            
\usepackage{booktabs}       
\usepackage{amsfonts}       
\usepackage{nicefrac}       
\usepackage{microtype}      

\usepackage{microtype}
\usepackage{graphicx}
\usepackage{subfigure}
\usepackage{booktabs} 
\usepackage{amsfonts}
\usepackage{amsmath}
\usepackage{multirow}
\usepackage{algorithmic}
\usepackage[ruled]{algorithm2e}
\usepackage{wrapfig}
\usepackage{float}
\usepackage{tabularx}

\title{Learning Deep Generative Models with Annealed Importance Sampling}

\author{%
  Xinqiang Ding \thanks{xqding@umich.edu} \\
  Department of Neurobiology \\
  The University of Chicago \\
  Chicago, IL 60637 United States \\
  \And
  David J. Freedman  \\
  Department of Neurobiology and \\ The Grossman Institute for Neuroscience, Quantitative Biology and Human Behavior \\
  The University of Chicago \\
  Chicago, IL 60637 United States \\
}

\begin{document}

\maketitle

\begin{abstract}
Variational inference (VI) and Markov chain Monte Carlo (MCMC) are two main approximate approaches for learning deep generative models by maximizing marginal likelihood.
In this paper, we propose using annealed importance sampling, which is equivalent to the Jarzynski equality from non-equilibrium thermodynamics, for learning deep generative models.
Our proposed approach bridges VI with MCMC.
It generalizes VI methods such as variational auto-encoders and importance weighted auto-encoders (IWAE) and the MCMC method proposed in \cite{hoffman2017learning}.
It also provides insights into why running multiple short MCMC chains can help learning deep generative models.
Through experiments, we show that our approach yields better density models than IWAE and can effectively trade computation for model accuracy without increasing memory cost.
\end{abstract}

\section{Introduction}
Deep generative models with latent variables are powerful probabilistic models for high dimensional data.
One of the challenges for learning such models by maximizing marginal likelihood is sampling from posterior distributions of latent variables, because the posterior distributions are often intractable and  have complex dependency structures.
Two main approximate approaches for learning such models are variational inference (VI) \cite{wainwright2008graphical,blei2017variational} and Markov chain Monte Carlo (MCMC) \cite{neal1993probabilistic}.

Here let us assume the generative model of interest is defined by a joint distribution of observed data variables $x$ and continuous latent variables $z$: $p_\theta(x, z) = p_\theta(z) p_\theta(x|z)$, where $\theta$ represents parameters of the generative model.
Given training data $x$, we are interested in learning the generative model $p_\theta(x, z)$ by maximizing its marginal likelihood, i.e., maximizing $\log p_\theta(x) = \log \int{p_\theta(x,z)} dz$.
Because $\log p_\theta(x)$ is usually a high dimensional integration when $z$ is high-dimensional and has complex dependence structure between its components, computing $\log p_\theta(x)$ is expensive.
However, we note that maximizing $\log p_\theta(x)$ does not necessarily require computing the value of $\log p_\theta(x)$.
If we use first order optimization methods for training, what is necessarily required is the gradient of $\log p_\theta(x)$ with respect to $\theta$, i.e., 
\begin{equation}
\label{equ:gradient_target}
\nabla_\theta \log p_\theta(x) = \mathop{\mathbb{E}}_{z \sim p_\theta(z|x)}{\big[\nabla_\theta \log p_\theta(x,z)\big].}
\end{equation}
As shown in (\ref{equ:gradient_target}), computing $\nabla_\theta \log p_\theta(x)$ is equivalent to calculating the expectation of $\nabla_\theta \log p_\theta(x,z)$ with respect to the posterior distribution $p_\theta(z|x)$.
Because the posterior distribution $p_\theta(z|x)$ is usually analytically intractable and it is often even difficult to draw samples from it, accurately computing $\nabla_\theta \log p_\theta(x)$ based on Eq. (\ref{equ:gradient_target}) is computationally expensive.
Therefore, efficient approximation methods are required to estimate the expectation in (\ref{equ:gradient_target}).

Several VI methods, including variational auto-encoders (VAEs) \cite{kingma2013auto,rezende2014stochastic} and importance weighted auto-encoders (IWAEs) \cite{burda2015importance} can be viewed/understood as devising ways of approximating the expectation in (\ref{equ:gradient_target}).
VAEs use an amortized inference model $q_\phi(z|x)$ to approximate the posterior distribution $p_\theta(z|x)$.
Both the generative model and the inference model are learned by maximizing the evidence lower bound (ELBO):
\begin{equation}
\label{equ:elbo}
\mathcal{L}(\theta, \phi) = \mathop{\mathbb{E}}_{z \sim q_\phi(z|x)} [ \log p_\theta(x,z) - \log q_\phi(z|x) ].
\end{equation}
In maximizing ELBO, $\theta$ follows the gradient 
\begin{equation}
\label{equ:gradient_elbo}
\nabla_\theta \mathcal{L}(\theta, \phi) = \mathop{\mathbb{E}}_{z \sim q_\phi(z|x)} \big[\nabla_\theta \log p_\theta(x,z) \big].
\end{equation}
In VAEs, the expectation in (\ref{equ:gradient_elbo}) is usually estimated by drawing one sample from $q_\phi(z|x)$.
If we view VAEs as an approximate method for maximum likelihood learning, comparing (\ref{equ:gradient_elbo}) to (\ref{equ:gradient_target}), we note that VAEs can be viewed as using the expectation in (\ref{equ:gradient_elbo}) to approximate the expectation in (\ref{equ:gradient_target}) using one sample from $q_\phi(z|x)$.
The rationale behind the approximation is that the inference model $q_\phi(z|x)$ is optimized to approximate the posterior distribution $p_\theta(z|x)$.
However, when the inference model can not approximate the posterior distribution well due to either its limited expressibility or issues in optimizing the inference model, the estimator based on (\ref{equ:gradient_elbo}) will have large bias with respect to the target expectation in (\ref{equ:gradient_target}).
A canonical approach to reduce the bias in (\ref{equ:gradient_elbo}) is using multiple samples with importance sampling \cite{robert2013monte}.
Specifically, we can draw $K$ independent samples, $z_1, ..., z_K$, from $q_\phi(z|x)$.
Then the expectation in (\ref{equ:gradient_target}) is estimated using 
\begin{equation}
\label{equ:gradient_iwae}
\sum_{k=1}^{K} \widetilde{w_k} \nabla_\theta \log p_\theta(x,z_k)
\end{equation}
where $\widetilde{w_k} = w_k / \sum_{i=1}^{K}{w_i}$ and $w_k = w(x,z_k,\theta,\phi) = p_\theta(x,z_k) / q_\phi(z_k|x)$. 
The same gradient estimator as (\ref{equ:gradient_iwae}) was used in IWAEs, but in IWAEs the estimator was devised as the gradient of a new ELBO function based on multiple ($K$) samples \cite{burda2015importance}:
\begin{equation}
\label{equ:elbo_k}
\mathcal{L}_K(\theta, \phi) = \mathop{\mathbb{E}}_{z_1,...,z_K  \sim q_\phi(z|x)} \Bigg[ \log \frac{1}{K} \sum_{k=1}^{K}\frac{p_\theta(x,z_k)}{q_\phi(z_k|x)} \Bigg].
\end{equation}
It is easy to verify that the estimator in (\ref{equ:gradient_iwae}) is an unbiased estimator of $\nabla_{\theta} \mathcal{L}_K(\theta, \phi)$.
Again, if we view IWAEs as an approximate method for maximum likelihood learning, IWAEs can be viewed as using the estimator in (\ref{equ:gradient_iwae}) to approximate the expectation in (\ref{equ:gradient_target}).

MCMC approaches \cite{neal1993probabilistic} aim to sample from posterior distributions by running a Markov chain with posterior distributions as its stationary distributions.
Because it could take a large number of steps for a Markov chain to converge, MCMC approaches were known as much slower than VI and were not as widely used as VI approaches for learning deep generative models, especially on large datasets.
Compared with rapid developments of VI approaches in recent years, relatively fewer studies investigate the use of MCMC approaches for learning deep generative models \cite{hoffman2017learning,han2017alternating}.

\section{Method}
In this work, we propose using annealed importance sampling (AIS) \cite{neal2001annealed} (equivalent to the Jarzynski equality \cite{jarzynski1997nonequilibrium} from non-equilibrium thermodynamics) to estimate the expectation in (\ref{equ:gradient_target}).
As a generalization of importance sampling, AIS \cite{jarzynski1997nonequilibrium,neal2001annealed} uses samples from a sequence of distributions that bridge an initial tractable distribution, which is $q_\phi(z|x)$ in our case, with a final target distribution, which is $p_\theta(z|x) \propto p_\theta(x,z)$.
To bridge $q_\phi(z|x)$ with $p_\theta(z|x) \propto p_\theta(x, z)$, we construct a sequence of intermediate distributions whose probability densities are proportional to $f_1(z), ..., f_{T-1}(z)$ and 
\begin{equation}
\label{equ:ais_inter}
f_t(z) = f_0(z)^{1-\beta_t} f_T(z)^{\beta_t},
\end{equation}
where $f_0(z) = q_\phi(z|x), f_T(z) = p_\theta(x,z)$.
$\beta_t$ are inverse temperatures and satisfy the condition $0 = \beta_0 \leq ... \leq \beta_T = 1$.
To estimate the expectation in (\ref{equ:gradient_target}), we can generate $K$ samples $\{z_1, ..., z_K\}$ and compute their corresponding weights $\{w_1, ..., w_K\}$ as follows.
To generate the $k$th sample $z_k$ and calculate its weight $w_k$, a sequence of samples $\{z_k^0, ..., z_k^T\}$ is generated using the following procedure. 
Initially, $z_k^0$ is generated by sampling from the distribution $f_0(z) = q_\phi(z|x)$. 
Here $q_\phi(z|x)$ is learned by optimizing the ELBO (\ref{equ:elbo}) as in VAEs.
For $1 \le t \le T$, $z_k^t$ is generated using a reversible transition kernel $T_t(z|z_k^{t-1})$ that keeps $f_t$ invariant.
The transition kernel $T_t(z|z_k^{t-1})$ is constructed using the Hamiltonian Monte Carlo (HMC) sampling method \cite{neal2011mcmc} in which the potential energy function $U_t(z)$ is set to $U_t(z) = -\log f_t(z)$.
After $T$ steps the sample $z_k$ is set to $z_k = z_k^T$ and its weight $w_k$ is calculated as:
\begin{equation}
w_k = \frac{f_1(z^0)}{f_0(z^0)}\frac{f_2(z^1)}{f_1(z^1)} ... \frac{f_{T-1}(z^{T-2})}{f_{T-2}(z^{T-2})}\frac{f_T(z^{T-1})}{f_{T-1}(z^{T-1})}.
\end{equation}
With the generated $K$ samples ${z_1, ..., z_K}$ and their weights ${w_1, ..., w_K}$, the expectation in (\ref{equ:gradient_target}) is estimated using
\begin{equation}
\label{equ:ais_estimator}
\nabla_\theta \log p_\theta(x) = \mathop{\mathbb{E}}_{z \sim p_\theta(z|x)}{\big[\nabla_\theta \log p_\theta(x,z)\big]} \simeq \sum_{k=1}^{K}\widetilde{w_k} \nabla_\theta \log p_\theta(x,z_k),
\end{equation}
where $\widetilde{w_k} = w_k / \sum_{i=1}^{K}{w_i}$ are normalized weights.
In summary, the detailed procedures of our proposed method are described in Algorithm (\ref{algo:ais}).

\begin{algorithm}[h!]
\caption{Learning Deep Generative Models with Annealed Importance Sampling}
\label{algo:ais}
\begin{algorithmic}
\STATE \textbf{Require:}
\STATE $x$: training data 
\STATE $K$: the number of annealed importance weighted samples 
\STATE $p_\theta(x,z)$: the generative model 
\STATE $q_\phi(z|x)$: the inference model 
\STATE $T$: the number of inverse temperatures 
\STATE $\{\beta_t: 0=\beta_0 \leq ... \leq \beta_{T} = 1\}$: inverse temperatures for the generative model
\STATE $\{\epsilon_t: t = 1,...,T\}$: the step sizes used in leapfrog integration of HMC at each inverse temperature
\STATE $L$: the number of integration steps in HMC
\STATE \textbf{Calculate Gradients and Optimize Parameters:}
    \WHILE{ $\theta, \phi$ \textit{not converged}}
    \STATE sample example(s) $x$ from the training data;
    \STATE \textbf{update the generative model parameter $\theta$} 
    \STATE set $\log w_k = 0$ for $k = 1,...,K$; 
    \STATE sample $z^0 = [z^0_1, z^0_2, ..., z^0_K]$, where $z^0_k$ are i.i.d. samples from $q_\phi(z|x)$;
    \STATE $\log w_k \leftarrow (\beta_1 -\beta_{0}) [\log p_\theta(x,z^{0}_k) -\log q_\phi(z^{0}_k)]$ for $k = 1,...,K$;
    \FOR{$t \leftarrow 1$ {\bfseries to} $T-1$}
    \STATE $z^{t}$ = HMC$(z^{t-1}, \beta_{t}, \epsilon_{t}, L)$, where the potential energy function is: $U_t(z) = -\log f_t(z)$ and $f_t(z) = q_\phi(z|x)^{1-\beta_t} p_\theta(x,z)^{\beta_t}$;
    \STATE $\log w_k \leftarrow \log w_k + (\beta_t - \beta_{t-1}) [\log p_\theta(x,z^{t-1}_k) - \log q_\phi(z^{t-1}_k)]$ for $k = 1,...,K$;
    \ENDFOR
    \STATE $z^{T}$ = HMC$(z^{T-1}, \beta_{T}, \epsilon_{T}, L)$ or set $z^{T} = z^{T-1}$
    \STATE set $z = [z_1, ..., z_K] = [z_1^{T}, ..., z_K^{T}]$ and $\widetilde{w_k} = w_k / \sum_{i=1}^{K}{w_i}$; \\
    \STATE estimate the gradient $\nabla_\theta \log p_\theta(x)$ with $\delta_\theta = \sum_{k=1}^{K} \widetilde{w_k} \nabla_\theta \log p_\theta(x,z_k)$;\\
    \STATE apply gradient update to $\theta$ using $\delta_\theta$; \\
    \STATE \textbf{update the inference model parameter $\phi$} \\
    \STATE sample $\epsilon \sim \mathcal{N}(0, \textbf{I})$; \\
    \STATE set $z = \mu(\phi, x) + \sigma(\phi, x) \odot \epsilon $ and calculate $\mathcal{L}(\theta, \phi)$; \\
    \STATE estimate the gradient $\delta_\phi = \nabla_\phi \mathcal{L}(\theta, \phi)$ with the reparameterization trick; \\
    \STATE apply gradient update to $\phi$ using $\delta_\phi$;
    \ENDWHILE
\end{algorithmic}
\end{algorithm}

Our approach (Algorithm \ref{algo:ais}) is most closely related to the IWAE approach \cite{burda2015importance} and Matthew D. Hoffman's HMC approach (MH-HMC) \cite{hoffman2017learning} for learning deep generative models with MCMC.
Both methods can be viewed as special cases of our proposed approach.
The IWAE approach corresponds to setting $T = 1$ and $z^{1} = z^{0}$ in Algorithm (\ref{algo:ais}).
The MH-HMC approach \cite{hoffman2017learning} is equivalent to setting $K = 1$  and $0 = \beta_0 < \beta_1 = ... = \beta_T = 1$.
Previous studies showed that IWAEs can also be interpreted as optimizing the standard ELBO (\ref{equ:elbo}) using a more complex variational distribution that is implicitly defined by importance sampling (IS) \cite{cremer2017reinterpreting,mnih2016variational}.
Similar interpretation can be used to understand Algorithm \ref{algo:ais}.
Specifically, when using (\ref{equ:gradient_iwae}) or (\ref{equ:ais_estimator}) to estimate the expectation in (\ref{equ:gradient_target}), IS or AIS implicitly defines a proposal distribution, $q_{\text{IS}}(z|x)$ or $q_{\text{AIS}}(z|x)$, using samples from $q_\phi(z|x)$ to approximate the posterior $p_\theta(z|x)$.
We can sample from the implicitly defined proposal distributions, $q_{\text{IS}}(z|x)$ or $q_{\text{AIS}}(z|x)$, with Algorithm \ref{algo:implicit}.

\begin{minipage}{\linewidth}
\centering
\begin{minipage}{0.50\linewidth}
\begin{algorithm}[H]
\caption{Sampling $q_{\text{IS}}(z|x)$ or $q_{\text{AIS}}(z|x)$}
\label{algo:implicit}
\begin{algorithmic}
\STATE K: number of importance samples
\STATE L: number of integration steps in HMC
\STATE T: num of intermediate distributions
\STATE \textbf{case 1}: \textit{when importance sampling is used}
\STATE \hspace{4pt} sample $z_1, ..., z_K$ $\sim$ $q_\phi(z|x)$
\STATE \hspace{4pt} set $w_k = \frac{p_\theta(x, z_k)}{q_\phi(z_k|x)}$ and $\widetilde{w_k} = w_k/\sum_{k=1}^{K}{w_k}$
\STATE \textbf{case 2}: \textit{when AIS is used}
\STATE \hspace{4pt} sample $z_1, ..., z_K$ and compute $\widetilde{w_1}, ..., \widetilde{w_K}$ 
\STATE \hspace{4pt} with Algorithm \ref{algo:ais}
\STATE sample $j \sim$ \textit{Categorical}($\widetilde{w_1}, ..., \widetilde{w_K}$)
\STATE \textbf{return} $z_j$
\end{algorithmic}
\end{algorithm}
\end{minipage}
\hspace{0.05\linewidth}
\begin{minipage}{0.4\linewidth}
\begin{figure}[H]
\centerline{\includegraphics[width=\columnwidth]{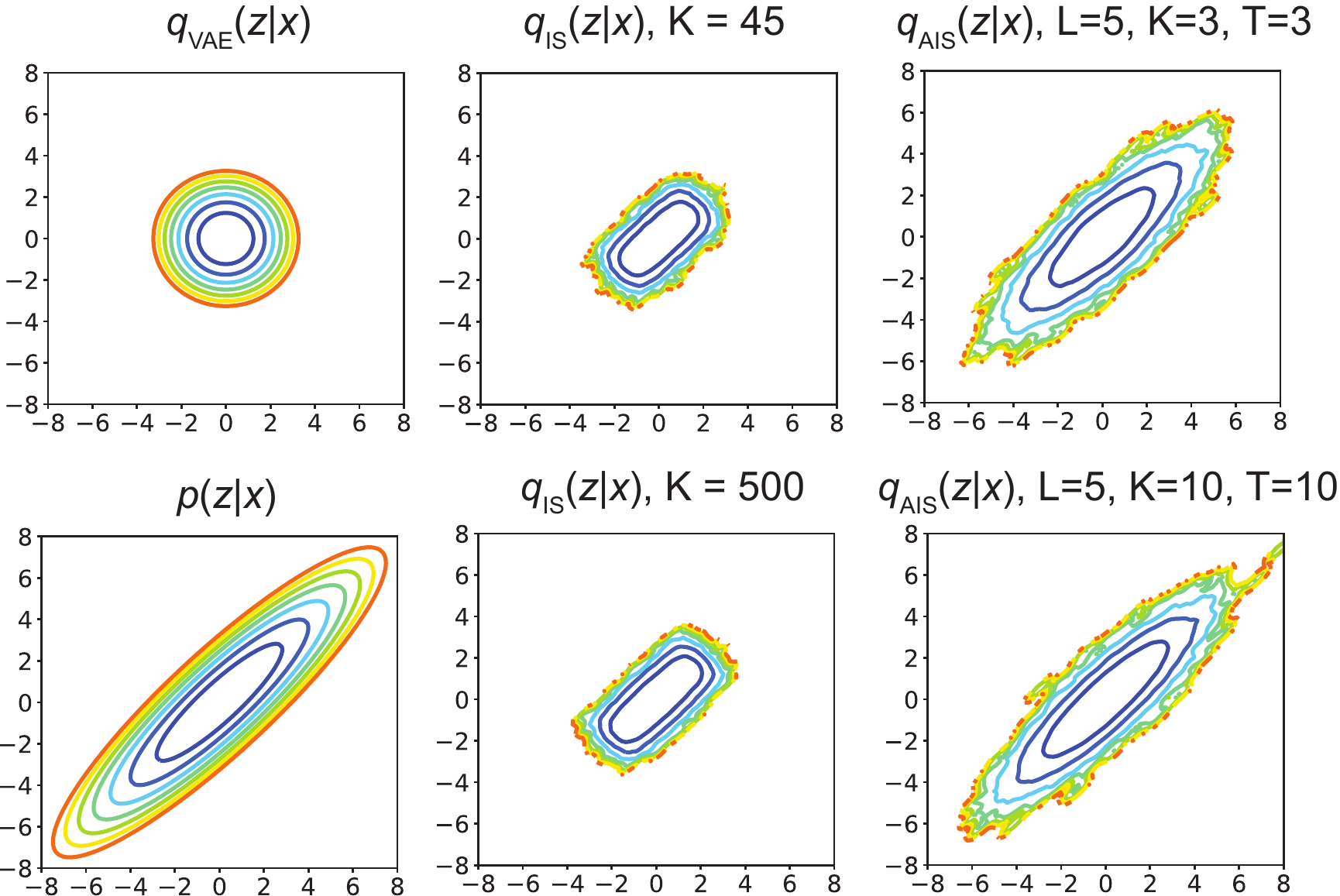}}
\caption{Comparison of implicit distributions $q_{\text{IS}}(z|x)$ and $q_{\text{AIS}}(z|x)$ for approximating the target distribution $p(z|x)$.}
\label{fig}
\end{figure}
\end{minipage}

\end{minipage}

One way to compare IWAEs and Algorithm (\ref{algo:ais}) is to compare the computational efficiency of $q_{\text{IS}}(z|x)$ and $q_{\text{AIS}}(z|x)$ for approximating the posterior $p_\theta(z|x)$.
To do that, we apply them in the following simple example.
The target distribution $p(z|x)$ is chosen to be a normal distribution of two correlated random variables.
The proposal distribution $q(z|x)$ is set to the normal distribution of two independent random variables that minimizes the Kullback-Leibler (KL) divergence from $q(z|x)$ to $p(z|x)$.
The computational cost of IS using $q_{\text{IS}}(z|x)$ increases linearly with $K$ and the cost of AIS using $q_{\text{AIS}}(z|x)$ scales linearly with  $L \times K \times T$. 
To make a fair comparison, we compare $q_{\text{IS}}(z|x)$ and $q_{\text{AIS}}(z|x)$ under the same computational cost, i.e., $K$ in IS is equal to $L \times K \times T$ in AIS.
The inverse temperatures $\beta_t$ in AIS are set to changing linearly with $t$ from 0 to $T$.
The results are shown in Fig. (\ref{fig}). 
Both $q_{\text{IS}}(z|x)$ and $q_{\text{AIS}}(z|x)$ become better approximation of $p(z|x)$ when increasing $K$ or $L \times K \times T$.
With the same amount of computational cost, $q_{\text{AIS}}(z|x)$ approximates the target distribution better than $q_{\text{IS}}(z|x)$.
The better performance of $q_{\text{AIS}}(z|x)$ for approximating $p(z|x)$ is expected to help Algorithm (\ref{algo:ais}) learn better generative models than IWAEs.

\section{Experiment Results}
\begin{table*}[h]
\caption{Results of IWAE-DReG and our approach on the Omniglot and the MNIST dataset.}
\label{table:omniglot}
\begin{center}
\begin{small}
\begin{tabular}{lcccccc}
\toprule
\multirow{2}{*}{$\approx \log p(x)$}& \multicolumn{3}{c}{Omniglot} &  \multicolumn{3}{c}{MNIST} \\
\cmidrule(r){2-4} \cmidrule(r){5-7}  
 & K = 1  & K=5 & K=50 & K = 1  & K=5 & K=50 \\
\midrule
IWAE-DReG      & -109.41 & -106.11 & -103.91 & -86.90 & -85.52 & -84.38 \\ 
Ours (T = 5)   & -103.22 & -102.47 & -102.03 & -84.56 & -84.25 & -83.93 \\ 
Ours (T = 11)  & -102.45 & -101.94 & -101.64 & -84.14 & -83.78 & -83.63\\
\bottomrule
\end{tabular}
\end{small}
\end{center}
\end{table*}
\begin{table}[h]
\caption{Results of IWAE-DReG, MH-HMC and our approach on the Omniglot dataset with same computational cost.}
\label{table:eql}
\begin{minipage}{0.50\linewidth}
\resizebox{\columnwidth}{!}{
\begin{tabular}{lcc}
\toprule
& $\approx \log p(x)$ \\
\midrule
IWAE-DReG (K = 55) & -103.85\\
Ours (L = 5, K = 1, T = 11) &  -102.45\\
\midrule
IWAE-DReG (K = 275) & -103.13 \\
Ours (L = 5, K = 5, T = 11) &  -101.94\\
\midrule
IWAE-DReG (K = 2750) & -102.40 \\
Ours (L = 5, K = 50, T = 11) &  -101.64\\
\bottomrule
\end{tabular}
}
\end{minipage}
\begin{minipage}{0.45\linewidth}
\resizebox{\columnwidth}{!}{
\begin{tabular}{lc}
\toprule
& $\approx \log p(x)$ \\
\midrule
MH-HMC (L = 5, K = 5, T = 5) & -102.57 \\
Ours (L = 5, K = 5, T = 5) & -102.47 \\
\midrule
MH-HMC (L = 5, K = 5, T = 11) & -101.32 \\
Ours (L = 5, K = 5, T = 11) & -101.94 \\
\midrule
MH-HMC (L = 5, K = 50, T = 5) & -102.32 \\
Ours (L = 5, K = 50, T = 5) &  -102.03 \\
\midrule
MH-HMC (L = 5, K = 50, T = 11) & -101.25 \\
Ours (L = 5, K = 50, T = 11) &  -101.64 \\
\bottomrule
\end{tabular}
}
\end{minipage}
\end{table}

We conducted a series of experiments to evaluate the performance of our proposed algorithm (\ref{algo:ais}) on learning deep generative models using the Omniglot \cite{lake2015human} and the MNIST \cite{lecun1998gradient} datasets.
We used same generative models and same inference models as that used in the IWAE study \cite{burda2015importance}.
Same models are also learned using the two closely related approaches: IWAEs and  MH-HMC \cite{hoffman2017learning}.
Because previous study \cite{tucker2018doubly} showed that IWAEs with doubly reparameterized gradient estimators (IWAE-DReG) can improve its performance, we used IWAE-DReG in all computations involving IWAE.
Following \cite{wu2016quantitative,hoffman2017learning,cremer2018inference}, we evaluate learned generative models using marginal likelihoods estimated with AIS \cite{neal2001annealed} on test datasets.
To be confident that the estimated likelihoods are accurate enough for comparing models, we follow \cite{wu2016quantitative} to empirically validate the estimates using Bidirectional Monte Carlo \cite{grosse2015sandwiching}.

For models trained with IWAE-DReG, the marginal likelihood increases on both  datasets when the value of $K \in \{1,5,50\}$ increases.
This agrees with previous studies \cite{burda2015importance,tucker2018doubly}.
For a fixed $K$, our approach with $T = 5$ or $T = 11$ produces better density models than IWAE-DReG.
If we view IWAE-DReG as a special case of our approach with $T = 1$, then the performance of our approach improves when increasing the value of either $K$ or $T$.
This is because increasing $K$ or $T$ can make implicit distributions defined by IS and AIS better approximate the posterior distribution, which in turn improves estimators in (\ref{equ:gradient_iwae}) and (\ref{equ:ais_estimator}) for estimating the expectation in (\ref{equ:gradient_target}).
For a fixed $K$, the computational cost of our approach is about $L \times T$ times that of IWAE-DReG. 
Therefore results in Table \ref{table:omniglot} only show that our approach is an effective way of trading computation for model accuracy.
To show that our approach is also computationally more efficient, we also compared learned models trained using IWAE-DReG and our approach with the same computational cost on the Omniglot dataset.
The result is shown in Table \ref{table:eql} (left).
It shows that, with the same computational cost, our approach leads to better density models than IWAE-DReG.

We also compared our approach with the MH-HMC method for learning deep generative models with same computational cost.
The MH-HMC method \cite{hoffman2017learning} always sets $\beta_t$ as $0 = \beta_0 < \beta_1 = ... = \beta_T = 1$.
In our approach, $\beta_t$ are free to choose as long as they satisfy the constraints that $0 = \beta_0 \leq \beta_1 \leq ... \leq \beta_T = 1$.
In this study, we set $\beta_t$ to change linearly between 0 and 1.
When $K = 1$, the MH-HMC approach is a special case of our approach for setting  $\beta_t$.
When $K > 1$, the MH-HMC method is different from our approach in the way of weighting samples.
The marginal likelihoods of models learned using the MH-HMC method and our approach are shown in Table \ref{table:eql} (right).
Similar to models trained with our approach, models trained with the MH-HMC method also improves when increasing the value of $K$ or $T$.
For the cases where $T = 5$, our approach leads to better density models, whereas for the cases where $T = 11$, the MH-HMC method leads to better density models.

\section*{Broader Impact}
The algorithm developed in this work can be applied to improve the training of various generative models for applications including image recognition/generation and nature language processing. 
It can effectively trade computational cost for model accuracy without increasing memory cost.
With available computational resources rapidly increasing, the algorithm can greatly benefit researchers in various scientific fields that are in need of training generative models.

We note that the algorithm is not designed to be able to remove biases that may exist in the training data. Models trained with the algorithm on biased datasets may inherit those biases such as gender, racial, nationality or age biases in the datasets. Therefore, we encourage users of this algorithm to be careful of any bias that may exist in their datasets. If there is any bias found in the dataset, they should either remove the bias before applying the algorithm or combine the algorithm with other approaches that can eliminate the effect of bias in datasets on trained models.

%

\begin{ack}
Funding in direct support of this work: Vannevar Bush Faculty Fellowship from USA Department of Defense
\end{ack}

\bibliographystyle{unsrt}
\bibliography{references}

\newpage

\end{document}